# Data-Driven Prediction of Maternal Nutritional Status in Ethiopia Using Ensemble Machine Learning Models

Amsalu Tessema[1]    Tizazu Bayih[1]    Kassahun Azezew[2]    Ayenew Kassie[3]
Department of Software Engineering, College of Engineering and Technology, Injibara University, Ethiopia[1]
Department of Computer Science, College of Engineering and Technology, Injibara University, Ethiopia[2]
Department of Information Technology, College of Engineering and Technology, Injibara University, Ethiopia[3]
amsalu.tessema@inu.edu.et[1] , tizazu.bayih@inu.edu.et[1]  azeze2912@gmail.com[2]  ayenew.kassie@inu.edu.et[3]

*Abstract*— Malnutrition among pregnant women poses a critical challenge to public health in Ethiopia, affecting maternal well-being and increasing the risk of adverse pregnancy outcomes. Despite significant efforts, traditional statistical methods used in nutritional assessments often fall short in capturing complex, multidimensional factors that influence nutritional status. The study aims to develop a predictive model using ensemble machine learning techniques to assess the nutritional status of pregnant women in Ethiopia, leveraging data from the Ethiopian demographic and health survey, managed by the central statistics agency. The research uses a dataset comprising 18,108 records spanning five survey years (2005–2020), with 30 attributes reflecting key socio-demographic and health indicators. A rigorous data preprocessing phase was undertaken, including handling missing values, normalization, and data balancing using the SMOTE technique. Feature selection was conducted using ensemble methods to identify the most influential predictors. The study employs multiple supervised ensemble machine learning algorithms, including XGBoost, Random Forest, CatBoost, and AdaBoost, to develop and compare classification models. Among the tested models, Random Forest model delivered the highest performance and robustness, successfully classifying nutritional status into four categories normal, moderate malnutrition, severe malnutrition, and overnutrition with an accuracy of 97.87%, precision of 97.88%, recall of 97.87%, F1-score of 97.87%, and an outstanding ROC AUC of 99.86%. The results underscore the feasibility of ensemble machine learning in capturing nuanced patterns from complex datasets, offering timely insights for targeted interventions. The findings offer valuable implications for healthcare professionals, policymakers, and researchers by enabling early detection of nutritional risks and informing data-driven strategies to improve maternal health outcomes across Ethiopia.

*Keywords— Nutritional Status, Pregnant Women, Ensemble Machine Learning, Ethiopian Demographic and Health Survey EDHS, Random Forest, Ethiopia*

## I. NTRODUCTION

Nutrition is the general term for the manner in which organisms obtain and process the nutrients or minerals to sustain themselves and growth [1]. Nutrition in a mother is needed most in keeping her healthy and for proper growth and development of the baby. A mother's nutritional condition during pregnancy from its onset until lactation plays a significant role in shaping her own and that of her child's health [2]. One most important thing to mention would be, it is vitally compulsory for taking nutrients during pregnancy to meet the growing fetus's needs and requirements and that of the mother's needs and wants. [3]. Every woman needs proper nutrition to adequately perform their crucial roles in reproduction and child-rearing. [4].

When it comes to maternal nutrition, two main kinds of problems it addresses are. One is undernutrition in the form of micronutrient deficiencies (inadequate vitamins and minerals), stunting (short height for age), wasting (low weight for height), and underweight (low weight for age) [5]. The second is consuming more than required, wherein one is either obese or overweight and has diet related non-communicable diseases like diabetes, cancer, heart disease, or stroke [6]. Undernutrition is nevertheless an important public health issue in Ethiopia, even though the country still lacks a special nutrition policy. Poor nutrition during maternity leads to critical birth issues like low birth weight, preterm delivery, and stunted growth of the fetus [7]. On the other hand, healthy mothers provide infants a much better beginning in life [8]. Pregnancy is perhaps the most nutritionally demanding period that a woman will experience. The body's demands skyrocket: energy demands increase, new tissues are formed, and organs like the breasts and uterus are developed all magnifying the risk of malnutrition if diets are poor short [9]-[10].

Pregnant women are predominantly susceptible to malnutrition, which is a critical global public health issue with maternal disease and poor pregnancy outcomes. This is particularly significant among young pregnant women between the ages of 18-30 years in low socio-demographic index regions, where the epidemiological burden of malnutrition disorders is significantly greater [11]. Women are twice at risk of malnutrition due to high nutritional demands for pregnancy and lactation, and it carries a double burden of "dysnutrition," both stunting or micronutrient malnutrition and obesity or other chronic diseases related to nutrition. Women in reproductive age are most at risk of malnutrition both biologically and socially [12].

Malnutrition, or an imbalance of nutrients in consumption, is a very dangerous threat to women's health, especially during pregnancy. Malnutrition is the cause of such alarming outcomes as prenatal and neonatal death, low birth weight, stillbirth, and miscarriage. Root causes for malnutrition are also multi-factorial with influences according to socio-economic status, education level, cultural beliefs, and health care access, differing by geography regions [13]-[14]. In pregnant women, with inadequate nutrition, direct correlation has been established with greater prenatal and neonatal mortality from fetal compromise during intrauterine development. Furthermore, inadequate consumption of essential nutrients may lead to low birth weight, a condition linked with elevated infant mortality and lifetime health issues [15].

This situation is particularly prevalent amongst rural Ethiopian pregnant women, where the majority of them are starving due to socio-demographic factors such as education level, residence, household economic status, occupation, marital status, and wealth index [16]. Although neither patients nor clinicians fully understand the multifactorial causes of malnutrition, these factors have a central role in explaining the long-term nutritional issue amongst pregnant women. Against these complexities, health agencies and policymakers must create proper assessment of pregnant women's nutritional status to make effective nutrition-related plans and interventions to improve the health outcomes of the mother.

Pregnant women's malnutrition needs to be addressed through a multifaceted policy considering the various socio-economic, cultural, and health-related factors contributing to poor nutrition. Combing nutrition status into normal, severe, moderate, or over-nourished is surprisingly challenging even for health policymakers and decision-makers analyzing socioeconomic and demographic information. Researchers across the globe have researched the diet of pregnant women in great detail to evaluate nutritional status [17]-[18] and address issues regarding them. But here's the twist: almost all of these studies have been confined to descriptive statistics and cross-sectional analyses of local clinic data, usually for one town or city. These studies do address such key issues as socioeconomic status and demographics but at a price. The limitation is that, they can't prove cause-and-effect - they merely take a snapshot of now without telling us what will occur in the future [19].

To address these gaps, the paper applied machine learning approaches to build a predictive model for the identification of the nutrition status of pregnant women to identify potential risk factors and forecast varying levels of nutrition status, presenting policymakers and healthcare professionals with a robust instrument for planning successful prevention interventions and strategies.

This study proposes to bridge these research gaps by developing a sophisticated predictive framework using ensemble machine learning to classify the nutritional status of pregnant women in Ethiopia into four distinct categories: normal, severe malnutrition, moderate malnutrition, and overnutrition. By leveraging a large-scale, nationally representative dataset, the research aims to provide a robust and generalizable tool for maternal health risk assessment.

But it's not just about food. A blend of socioeconomic barriers, dietary habits, and health care access decides whether pregnant ladies get the nutrients they need [20], Early risk person identification is crucial for proactive intervention. Traditional methods, however, typically use manual data collection, which can be time intensive, error-prone, and difficult to scale. With advances in artificial intelligence and data-driven techniques, machine learning has become a crucial tool to make accurate predictions in the health sector. ML algorithms can process large datasets, identify hidden patterns, and make accurate predictions and are therefore very effective in ascertaining nutritional status. In the case of Ethiopia, applying ML for nutrition prediction would enhance early detection and facilitate better maternal healthcare planning. Despite the growing interest in applying ML in medicine, little evidence of its application in maternal nutrition screening, particularly in resource-poor settings like Ethiopia, exists. Machine learning attempts to learn by forming a set of hypotheses from the training data and combining them to build a model of predictions.

Ensemble learning has emerged as a powerful approach in supervised machine learning, significantly enhancing prediction accuracy and reliability by combining multiple models rather than depending on a single one [12]. Unlike conventional methods that rely on just one hypothesis, ensemble techniques integrate multiple hypotheses to achieve superior overall performance. Key methods include bagging such as Random Forest, which reduces variance by averaging predictions from models trained on different data subsets; boosting algorithms like AdaBoost and XGBoost that iteratively correct errors to reduce bias and strengthen weak learners; and stacking, which leverages a meta-learner to combine diverse models for even more precise predictions [13], [14], [15]. Ensembles can be homogeneous (using the same underlying algorithm) or heterogeneous (mixing different algorithms), but both strategies consistently outperform single-model approaches. At its core, ensemble learning operates on the principle that a group of "weak" learners can collectively surpass the performance of a single "strong" model, dramatically cutting variance and boosting accuracy beyond what any individual classifier could achieve [16]. This makes ensemble learning a cornerstone of modern machine learning.

## II. RELATED WORKS

Several studies have explored the use of ML in maternal health. Begum et al. [21] used Random Forest to predict the nutritional status of pregnant women in Bangladesh using BDHS data, achieving an accuracy of 74.75%. Dejene et al. [22] applied a homogeneous ensemble of CatBoost to predict anemia levels among Ethiopian PW with 97.6% accuracy. Khadidos et al. [23] proposed a quad-ensemble framework for maternal health risk classification, with Gradient Boosted Trees achieving 90% accuracy for high-risk cases.

A. Tesfaye and G. Sisay, R. 2022, proposed a ML approach to identifying under-nutrition and associated factors among pregnant women in public health care hospitals of Gedeo Zone, Southern Ethiopia: A cross-sectional study. The study found a 21% prevalence of undernutrition among pregnant women. They applied Logistic regression analysis on cross-sectional data, MUAC measurement, and socio-economic survey methods. However, the study lacks BMI inclusion and longitudinal data, which limits its ability to establish causality. Future researchers could benefit from incorporating machine learning-based predictive models to enhance early intervention strategies [24].

The work done by [25] discussed the determinants of pregnant women's dietary diversity practices in the Gurage zone, Ethiopia, following a community-based cross-sectional study. The approach has attempted a cross-sectional survey, qualitative interviews, and focus group discussions. In this, study only 42.1% of pregnant women followed an adequate dietary diversity. Factors like maternal education, women's empowerment, food security, and household wealth are significant. College-educated women are 3 more likely to follow diverse diets. The challenge here the cross-sectional nature and recall bias limit causal inference. Further research should use machine learning to model predictors of dietary diversity.

From those literature reviews, we have seen some approaches how they are applied different techniques to come up the solution to the research gaps However, these studies often focus on a specific aspect of nutrition (e.g., just anemia) or a binary risk classification. Our work distinguishes itself by performing a comprehensive four-class nutritional status prediction using a larger, multi-year Ethiopian dataset and providing a direct comparison of multiple state-of-the-art ensemble algorithms.

## III. METHODOLOGY FOR PREDICTING THE NUTRITIONAL STATUS OF PREGNANT WOMEN

The proposed model which is shown on Fig contains the overall research methodology. It addresses crucial stages such as data acquisition, preparation, feature extraction, train-test split, rule formation, risk factor determination, and artifact development, all organized based on the model structure developed. Each component of the model is described in subsequent sub-sections.

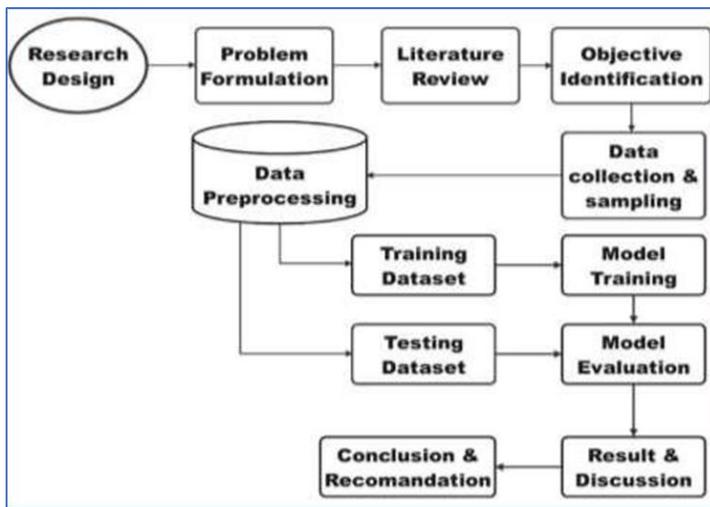

*Fig 1: illustrates the end-to-end research design implemented in this study.*

### A. Research Design

The study follows an experimental research design employing a mixed research approach that integrates qualitative and quantitative data collection and analysis techniques. The primary goal is to develop a machine learning model to predict the nutritional status of pregnant women in Ethiopia using data from the EDHS.

### B. problem formulation

The research process begins with problem formulation, where the research problem is defined, emphasizing the need for accurate and automated prediction methods to enhance maternal health. A careful literature review is then conducted to search for available methodologies, models, and research gaps areas and finally come up with specific research. The aim of this is to focus on data preparation and cleaning, developing predictive models, and their performance assessment using performance metrics.

### C. Data Source and Description

The study utilized secondary data from from the Ethiopian Demographic and Health Surveys (EDHS), which were collected and maintained by the Central Statistics Agency (CSA). This nationally representative dataset is a reliable source of information on maternal and child health indicators. The study aggregated data from four survey years: 2005, 2011, 2016, and 2019, resulting in an initial 18,108 records with 30 features. These attributes were precisely selected to encompass a wide range of factors influencing a woman's nutritional status and were grouped into five categories for clarity: Demographic (Age, region, residence, religion, marital status, children ever born), Educational (Mother's and husband's education, literacy), Socioeconomic (Wealth index, occupation, household assets), Health-related (Pregnancy status, contraceptive use, breastfeeding, BMI, anemia level, smoking), and Household/Environmental (Source of drinking water, type of toilet facility, cooking fuel type). The target variable, nutritional status, was derived from Body Mass Index (BMI) and categorized per WHO guidelines [26] into four classes: 0-Severe Malnutrition,1-Moderate Malnutrition, 2-Normal, 3-Overnutrition.

### D. Data Preprocessing and Engineering

Data preprocessing is a crucial step in transforming raw data into a clean, structured format suitable for machine learning [27]. The raw dataset presented several challenges, including missing values, class imbalance, and features requiring transformation. A rigorous preprocessing pipeline was essential to prepare the data for modeling. *Data Cleaning* (Mode imputation was applied to categorical variables, and mean imputation to numerical variables. *Data Transformation* (Continuous variables like Body Mass Index (BMI) were discretized into the defined categories. Categorical variables were encoded numerically (ordinal encoding for ordered categories, nominal for unordered). *Data Balancing* (The original class distribution was highly imbalanced (66.7% Normal, 22.9% Moderate, 8.8% Severe, 1.6% Overnutrition). The Synthetic Minority Over-Sampling Technique (SMOTE) was applied to create a balanced dataset of 44,164 instances, ensuring the model would not be biased toward the majority class.

*E. Feature Selection*

To enhance model performance and interpretability, feature selection was conducted. We employed a combination of: *Filter Methods*: Mutual Information, Chi-Square, and ANOVA F-test were used to rank features based on statistical properties. *Wrapper Method*: Sequential Backward Selection with a Random Forest estimator was used to find the optimal subset of features that maximized predictive accuracy. This process, combined with domain expertise, refined the feature set from 30 to 19 of the most predictive attributes [28].

*F. Model Development and Evaluation*

We selected four powerful ensemble learning algorithms for model development and comparison: *Random Forest (RF)*: A bagging algorithm that constructs a multitude of decision trees for robust and accurate classification. *XGBoost*: An optimized gradient boosting algorithm known for its speed and performance. *CatBoost*: A gradient boosting algorithm effective at handling categorical features with minimal preprocessing. *AdaBoost*: A boosting algorithm that focuses on correcting errors from previous learners.

*G. Train-Test Split the dataset*

Overfitting is a behavior that significantly reduces the accuracy of the model in creating good predictions on future data or on data not resembling the training set [29]-[30]. Therefore, the model is not highly effective when presented with real-life situations, which are heterogenous and varied. The necessity to choose an optimal train-test split cannot be overestimated. Optimal split should ideally be in proportion so that the model has sufficient data to learn well but also keeps sufficient data to test its performance on new data. A carefully chosen split reduces overfitting and bias, increases the model's capacity to generalize, and enables performance estimation. Correct partitioning also helps in hyperparameter tuning, uncertainty estimation, and cross-model or cross-algorithm comparison on an equal basis, ultimately making the model more trustworthy.

Stratified train-test splits of 70/30, 75/25, and 80/20 were experimented with; 80/20 split was chosen for optimal performance. 10-fold cross-validation was used for stable performance estimates. Model hyperparameter tuning was performed using Grid Search for parameters such as number of estimators, tree depth, and learning rates. Evaluation metrics included accuracy, precision, recall, F1-score, and ROC AUC.

## IV. EVALUATION RESULTS AND DISCUSION

*A. Experimental Setup*

The balanced dataset was split into training and testing sets using an 80-20 ratio. Model hyperparameters were meticulously tuned using Grid Search with 10-fold cross-validation to prevent overfitting and ensure generalizability. All experiments were implemented in Python using libraries including Scikit-learn, XGBoost, CatBoost, and Imbalanced-learn on the Google Colab platform.

*B. Performance Metrics*

To optimize the performance of each model, a systematic hyperparameter tuning process was performed using GridSearchCV. The optimal parameters were identified by exploring a predefined range of values for each model, ensuring that the final configuration mitigates overfitting and enhances generalization. The final models were evaluated using a comprehensive set of metrics, including *Accuracy* (Overall proportion of correct predictions). *Precision* (proportion of true positives among all positive predictions). *Recall* (proportion of actual positives correctly identified). *F1-Score* (Harmonic mean of precision and recall). *ROC AUC* (Area under the receiver operating characteristic curve, measuring the model's ability to distinguish between classes).

*Table 1: Evaluation matrices*

| Metric | Formula | Description |
|---|---|---|
| Accuracy | $\frac{TP + TN}{Population + Sample\ Size}$ | Overall performance of the model |
| Precision | $\frac{TP}{TP + TN}$ | How the positive description is accurate |
| Recall | $\frac{TP}{TP + FN}$ | Coverage of actual positive samples |
| F1-Score | $2 * \frac{(Precision * Recall)}{(Precision + Recall)}$ | The harmonic means of precision and recall |

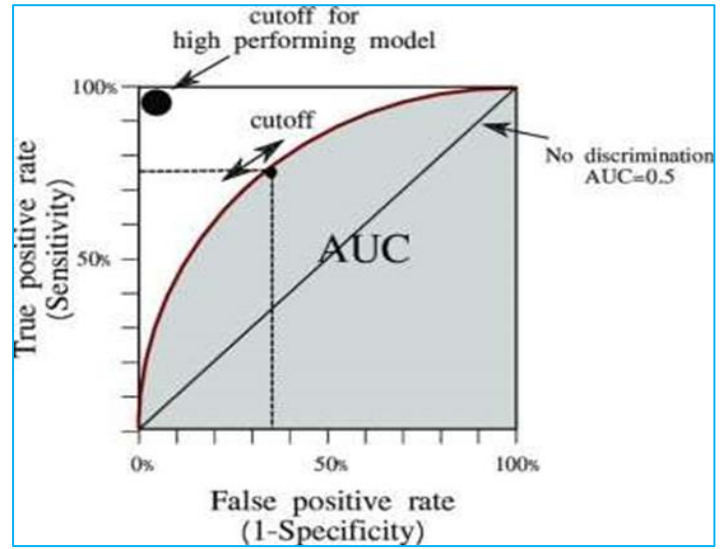

*Figure 2: Illustrates the end-to-end research design implemented in this study.*

The evaluation of the models' performance was a critical step in determining which algorithm would be most effective for the final predictive framework.

Table 2: Overall Performance of the Experiment with Different Train-Test Splits

| Split | Model | Accuracy (%) | Precision (%) | Recall (%) | F1 Score (%) | ROC AUC (%) |
|---|---|---|---|---|---|---|
| 70-30 | Random Forest | 97.47 | 97.47 | 97.47 | 97.47 | 99.84 |
| | AdaBoosting | 44.45 | 43.94 | 44.45 | 44.04 | 73.79 |
| | CatBoost | 90.60 | 90.55 | 90.60 | 90.54 | 98.44 |
| | XGBoost | 87.52 | 87.49 | 87.52 | 87.43 | 97.52 |
| 75-25 | Random Forest | 97.55 | 97.55 | 97.55 | 97.54 | 99.84 |
| | AdaBoosting | 45.90 | 45.99 | 45.90 | 45.93 | 74.09 |
| | CatBoost | 90.91 | 90.86 | 90.91 | 90.86 | 98.60 |
| | XGBoost | 87.37 | 87.33 | 87.37 | 87.30 | 97.60 |
| 80-20 | Random Forest | 97.78 | 97.79 | 97.78 | 97.78 | 99.87 |
| | AdaBoosting | 45.95 | 45.72 | 45.95 | 45.67 | 73.96 |
| | CatBoost | 91.15 | 91.10 | 91.15 | 91.10 | 98.66 |
| | XGBoost | 87.16 | 87.11 | 87.16 | 87.08 | 97.63 |

### C: Performance of Ensemble Models with using Hyperparameter Tuning

In machine learning, the performance of a model can largely be dictated by how the hyperparameters are set, so hyperparameter tuning is an essential process in building a model. To boost the precision of our ensemble models, Grid Search and Cross-Validation (GridSearchCV) was used to find the best set of parameters systematically. This technique systematically explores a predefined set of hyperparameter values and evaluates the model for every possible combination within the grid. For this study, GridSearchCV was applied to each ensemble classifier Random Forest, AdaBoost, CatBoost, and XGBoost across all train-test split experiments (70-30, 75-25, and 80-20). The optimal set of hyperparameters obtained through this process was then used to train the final models as provided in table 3. After hyperparameter tuning, the four ensemble models were evaluated on the held-out test set. Table 4 presents a comprehensive summary of their performance. along with their corresponding model configurations.

Table 3: Hyperparameter Tuning Results for Ensemble Algorithms

| Model | Key Hyperparameters |
|---|---|
| RF | n_estimators=400, max_depth=100, min_samples_split=3, max_features='sqrt', criterion='gini' |
| AdaBoost | n_estimators=100, base_estimator=DecisionTreeClassifier(criterion='entropy', max_depth=40, max_leaf_nodes=200, min_samples_split=3, max_features='sqrt') |
| XGBoost | n_estimators=500, max_depth=100, learning_rate=0.1, booster='gbtree', min_child_weight=1 |
| CatBoost | iterations=1000, depth=8, learning_rate=0.1, l2_leaf_reg=3, verbose=0 |

Table 4: Performance Comparison of Ensemble Models

| Model | Accuracy (%) | Precision (%) | Recall (%) | F1-Score (%) | ROC AUC (%) |
|---|---|---|---|---|---|
| **Random Forest** | **97.87** | **97.88** | **97.87** | **97.87** | 99.86 |
| XGBoost | 97.68 | 97.68 | 97.68 | 97.68 | **99.88** |
| CatBoost | 96.74 | 96.73 | 96.74 | 96.73 | 99.65 |
| AdaBoost | 90.73 | 90.80 | 90.73 | 90.75 | 96.94 |

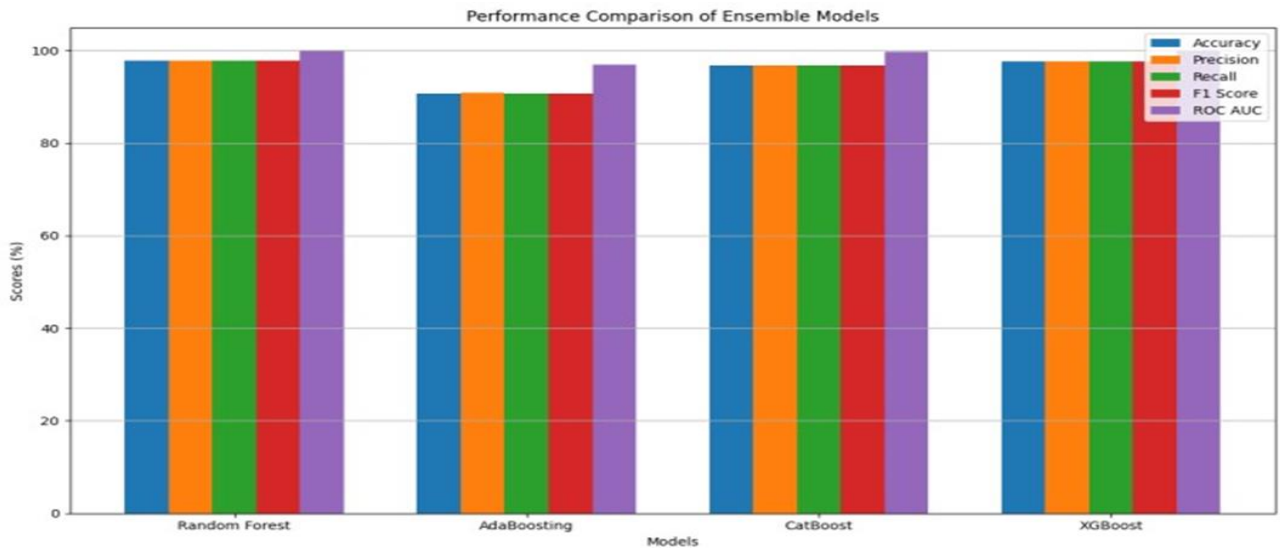

*Figure 3: The overall performance of the four experiments*

As we observe from both table and fig 3, the results clearly indicate that Random Forest is the best-performing model for this specific prediction task, achieving the highest scores across accuracy, precision, recall, and F1-score. XGBoost performed very closely, registering the highest ROC AUC score (99.88%), signifying an excellent ability to rank predictions correctly. CatBoost also demonstrated strong performance, while AdaBoost was significantly less accurate, suggesting it is less suited for the complexity of this multi-class problem.

### D: Confusion Matrix Analysis

The confusion matrices for each model (Fig. 4) provide detailed insight into class-wise performance. The diagonal dominance in all matrices confirms good predictive power. Random Forest and XGBoost show the least misclassification, particularly for the minority classes (Severe and Overnutrition), which is critical for a useful healthcare intervention tool.

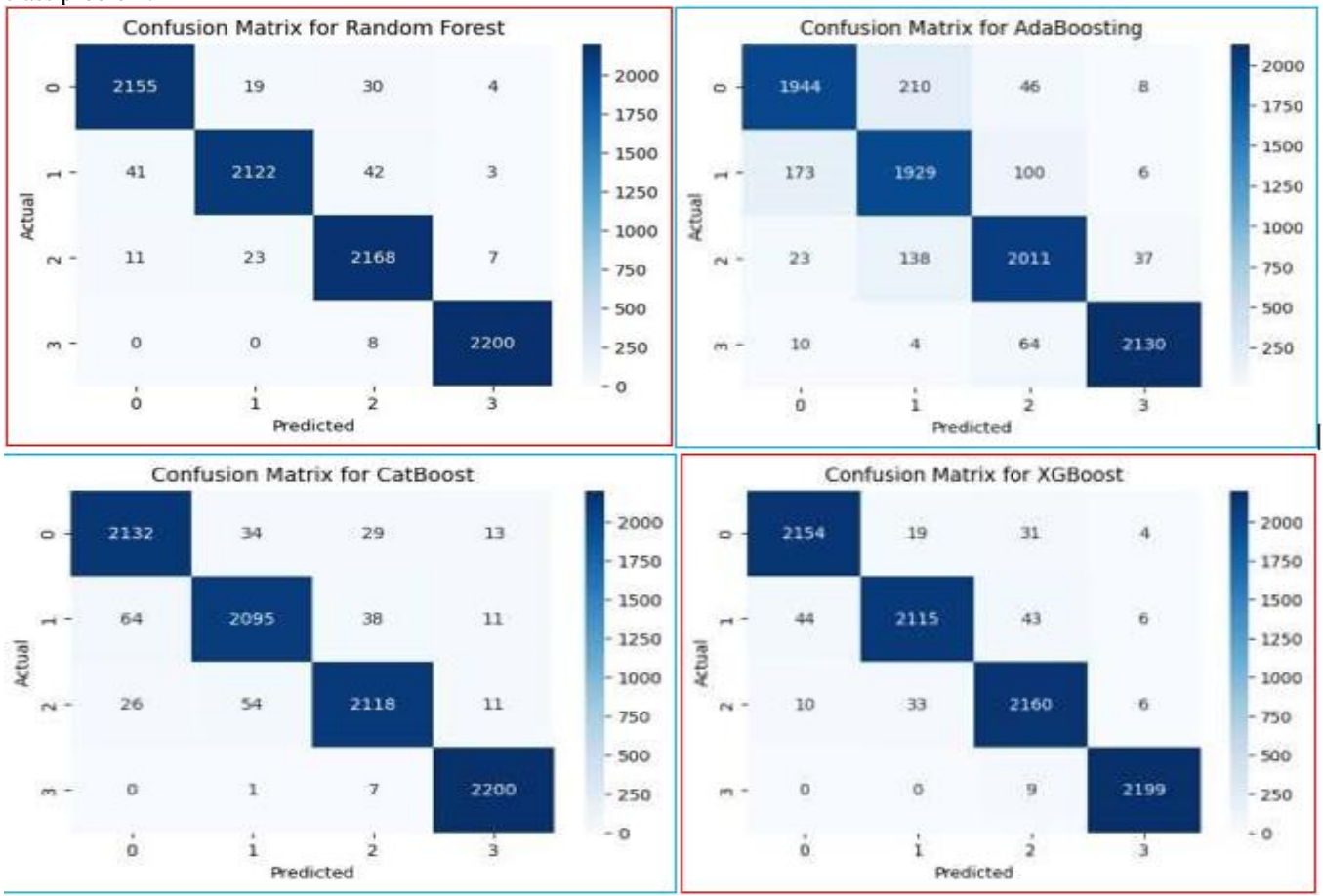

*Figure 4: Confusion Matrix - Random Forest, AdaBoost, Cat Boost, and XGBoost classifier*

As we have observed in the detailed analysis of the confusion matrices provides a more granular view of each model's performance on a class-by-class basis. The confusion matrix for the Random Forest model demonstrates its outstanding performance. The vast majority of predictions fall on the diagonal, indicating correct classifications. For instance, the model correctly classified 2,155 cases for class 0 (severe malnutrition), 2,122 for class 1 (moderate malnutrition), 2,168 for class 2 (normal), and 2,200 for class 3 (overnutrition). The high number of correct predictions for the severe malnutrition class is particularly significant from a public health perspective, as it confirms the model's reliability in identifying the most critical at-risk women. The confusion matrices for XGBoost and CatBoost show similar, albeit slightly less precise, patterns, while AdaBoost exhibits a higher degree of misclassification, particularly between adjacent classes.

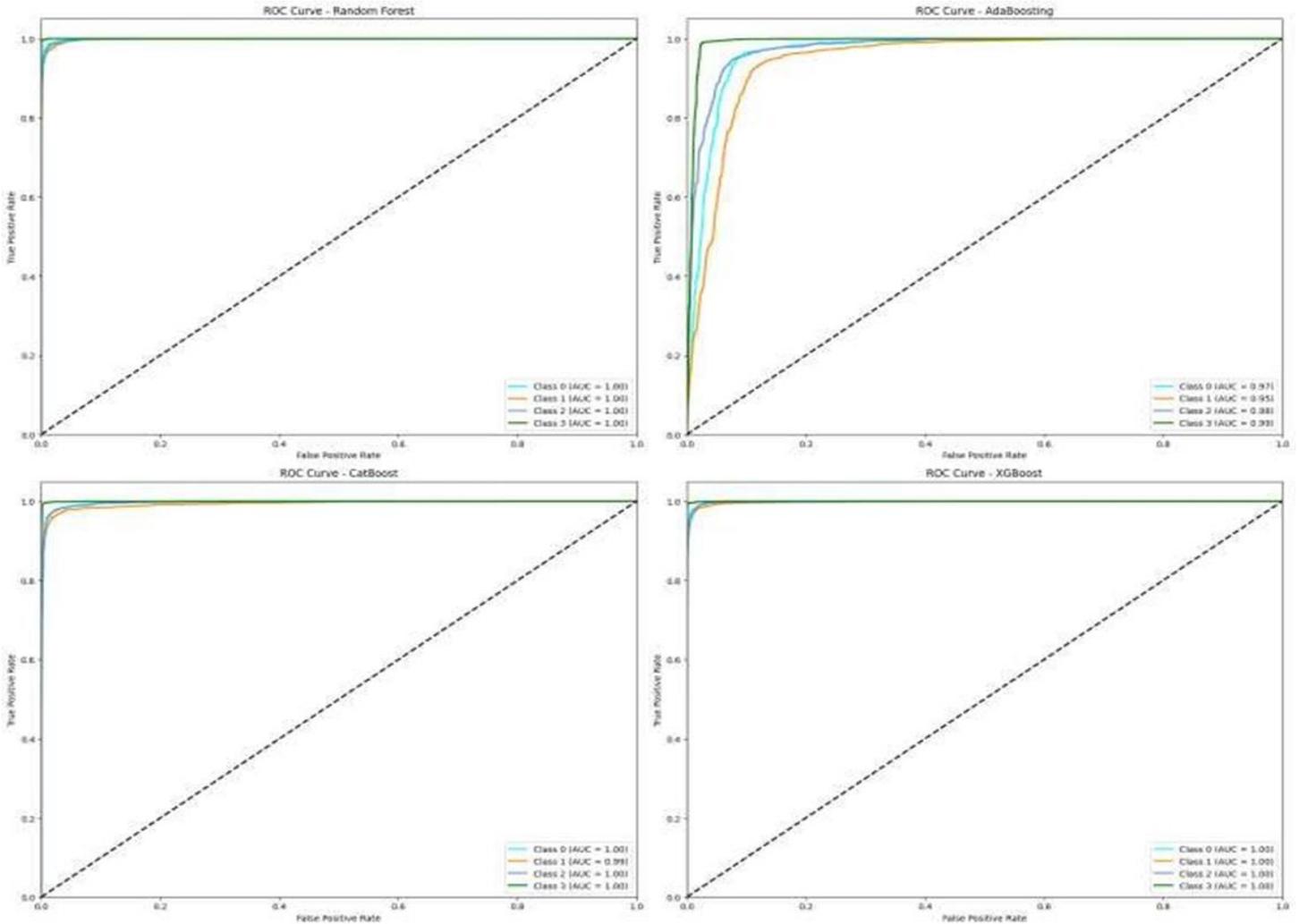

*Figure 5: Comparison of ROC Curve for Random Forest, AdaBoost, Cat Boost, and XGBoost classifier*

The ROC curves for the Random Forest model (Fig. 6) further validate its superior performance, with each class's curve hugging the top-left corner and AUC values very close to 1.00, indicating near-perfect classification ability for all four nutritional categories.

## VI. CONCLUSION AND FUTURE WORK

This research successfully developed a robust, multi-class predictive model for the nutritional status of pregnant women in Ethiopia using ensemble machine learning. By addressing significant data challenges such as class imbalance with SMOTE and leveraging a comprehensive, nationally representative dataset, the study has created a framework that surpasses the capabilities of traditional statistical and baseline machine learning methods.

The Random Forest model demonstrated exceptional performance, achieving an accuracy of 97.87%, a precision of 97.88%, a recall of 97.87%, and an F1-score of 97.87%. This high level of performance underscores the feasibility and effectiveness of using sophisticated machine learning algorithms to uncover nuanced patterns in complex public health data. Furthermore, the study's feature importance analysis provided critical insights by identifying the most influential predictors of maternal nutritional status. The findings indicate that factors such as Region, Mother's Age Group, Wealth Index, and Education Level are the key determinants, providing a data-driven mandate for policymakers to focus on improving systemic socioeconomic conditions in addition to direct nutritional interventions.

The contributions of this study are significantly provides a reproducible and generalizable framework for multi-class nutritional risk assessment, identifies specific, actionable risk factors for targeted interventions, and demonstrates the power of ensemble machine learning in a crucial public health domain. This work offers valuable implications for healthcare professionals, enabling the early detection of nutritional risks, and for policymakers, allowing for the formulation of evidence-based strategies to improve maternal health outcomes across Ethiopia.

*Recommendations for Future Research*

We suggest for the future researcher will consider: Incorporate clinical and biometric data to further enhance predictive accuracy and clinical relevance. Explore deep learning approaches to capture complex nonlinear relationships. Extend predictive modeling to other vulnerable populations such as lactating women and neonates. Develop integrated knowledge-based systems that automate diagnosis and intervention suggestions. Apply embedded feature selection methods like Lasso to streamline model complexity.